\documentclass[letterpaper, 10 pt, conference]{ieeeconf}  

\IEEEoverridecommandlockouts                              

\usepackage{cite}
\usepackage{float}
\usepackage{multicol}
\usepackage{amsmath,amssymb,amsfonts}
\usepackage{graphicx}
\usepackage{textcomp}
\usepackage{xcolor}
\usepackage[ruled,linesnumbered, vlined]{algorithm2e}
\DontPrintSemicolon
\SetCommentSty{myCommentStyle}

\usepackage{amsfonts}
\let\labelindent\relax
\usepackage{enumitem}
\usepackage{amsmath}
\usepackage{ragged2e}
\usepackage{graphicx}
\usepackage{subfig}
\usepackage[margin=0.792in]{geometry}
\usepackage{bbm}
\usepackage{algpseudocode} 
\usepackage{multirow}
\usepackage{xcolor}
\newcommand\Tstrut{\rule{0pt}{2.6ex}}         
\newcommand\Bstrut{\rule[-0.9ex]{0pt}{0pt}}   
\def\BibTeX{{\rm B\kern-.05em{\sc i\kern-.025em b}\kern-.08em
    T\kern-.1667em\lower.7ex\hbox{E}\kern-.125emX}}

\title{\LARGE \bf
Physics-informed Neural Time Fields for Prehensile Object Manipulation
}

\author{Hanwen Ren, Ruiqi Ni, and Ahmed H. Qureshi
\thanks{Hanwen Ren, Ruiqi Ni and Ahmed H. Qureshi are with the Department of Computer Science, Purdue University, West Lafayette, IN, USA, 47907. Email {\tt\small$\{$ren221, ni117, ahqureshi$\}@$purdue.edu}}%
}

\begin{document}

\maketitle
\thispagestyle{empty}
\pagestyle{empty} 
\begin{abstract}
Object manipulation skills are necessary for robots operating in various daily-life scenarios, ranging from warehouses to hospitals. They allow the robots to manipulate the given object to their desired arrangement in the cluttered environment. The existing approaches to solving object manipulations are either inefficient sampling based techniques, require expert demonstrations, or learn by trial and error, making them less ideal for practical scenarios. In this paper, we propose a novel, multimodal physics-informed neural network (PINN) for solving object manipulation tasks. Our approach efficiently learns to solve the Eikonal equation without expert data and finds object manipulation trajectories fast in complex, cluttered environments. Our method is multimodal as it also reactively replans the robot's grasps during manipulation to achieve the desired object poses. We demonstrate our approach in both simulation and real-world scenarios and compare it against state-of-the-art baseline methods. The results indicate that our approach is effective across various objects, has efficient training compared to previous learning-based methods, and demonstrates high performance in planning time, trajectory length, and success rates. Our demonstration videos can be found at https://youtu.be/FaQLkTV9knI.
\end{abstract}

\section{Introduction}
Prehensile Object Manipulation (POM) is one of the fundamental skills required for robots aiming to assist people in their daily lives \cite{billard2019trends}. In this skill, the robot is required to pick and place a given object from the start to the desired pose while avoiding collision with the surrounding environment. 

An ideal POM method should possess the following features. First, it should be able to learn diverse manipulation skills without needing expert demonstrations or trial-and-error-based interactions with the environment. Second, the learned manipulation skills should be generalizable to a wide range of objects in complex, cluttered settings. Third, the method should also leverage extrinsic dexterity to replan and re-grasp objects to reach the desired object pose. Lastly, the robot should find object manipulation trajectories quickly without significant computational load to allow for almost real-time executions.  

Despite widespread applications of POM, the existing methods lack one or more of the desired features mentioned above. Recent development led to physics-informed neural networks (PINN) \cite{raissi2019physics}, such as NTFields \cite{ni2022ntfields} and P-NTFields \cite{ni2023progressive}, which opened up new ways for motion planning. One of the great advantages of these methods is that they do not require expert training trajectories as they directly solve the Eikonal equation for motion generation. Also, the resulting trajectories often have a large safety margin towards the obstacles while maintaining a short Euclidean distance. Once trained, these methods can handle arbitrary start and goal configurations to find collision-free robot motion paths in a fraction of a second. However, these methods are currently only applied to standard motion planning tasks, and their applicability to POM is yet to be explored. In addition, there are a few limitations that need to be tackled in these PINN methods to meet the features of an ideal POM planner. First, the state-of-the-art PINN method can take a few hours to train, whereas an ideal method should converge faster for deployments. Second, the existing PINN approaches lack the generalization capability, which is required for the planner to handle diverse sets of objects.

Therefore, in this paper, we extend PINN to solve POM problems and meet the features of an ideal method. Our approach, \textbf{P}rehensile \textbf{O}bject \textbf{M}anipulating \textbf{Ne}ural \textbf{T}ime \textbf{F}ields (POM-NeTF), makes the following key contributions:
\begin{itemize}[leftmargin=*]
    \item A new formulation of the Eikonal equation based on Dirichlet energy minimization that allows for faster convergence of our physics-informed approach. 
    \item A novel neural architecture that enables PINN generalization to a wide range of objects.
    \item A novel POM method that allows the robot to leverage extrinsic dexterity to place objects at intermediate poses and replan their trajectory and grasp points to reach the goal.
    \item A unified framework that learns object manipulation skills without expert demonstrations and plans multi-step kinematically feasible object manipulation trajectories fast.
\end{itemize}
 
We tested our approach in both simulated and real-world scenarios and compared its performance against state-of-the-art PINN and classical methods. The results show that our approach converges quickly, finds viable paths with a high success rate, and requires very little computational time.

 \section{Related Work}
In this section, we present the relevant work solving the problem of POM. The prior work can be classified into three major categories, naming classical, imitation learning, and reinforcement learning methods.

The classical approaches use non-learning-based methods, such as sampling-based techniques that build graphs in the SE(3) domain and find an object manipulation trajectory \cite{huh2018constrained}. The resulting trajectory is then mapped back to robot configurations through inverse kinematics \cite{kucuk2006robot}. However, these methods are usually computationally expensive in cluttered environments, especially when optimizing for both collision avoidance and path length. There also exist methods \cite{cohen2010search} that define a set of robot motion primitives and search for a sequence of those primitives to manipulate the objects to their desired poses. Most of these methods also leverage extrinsic dexterity, i.e., leverage external contact points for object re-orientation and prehensile pushing. Despite the success, these methods lack generalization to novel scenarios and a large number of objects due to a fixed set of motion primitives \cite{hou2020manipulation, eppner2017visual}. 
The imitation learning methods learn object manipulation strategies from expert demonstrations\cite{gualtieri2018pick, song2020grasping, bain1995framework, ng2000algorithms}. One of the main advantages of imitation learning from demonstrations is that the robot can acquire new skills without explicit programming \cite{welschehold2016learning}. However, they are limited by their need for expert data, which is usually expensive to obtain. Another work \cite{simeonov2022neural} addresses this limitation by proposing SE(3)-equivariant neural descriptor fields (NDF) that, with just a few demonstrations, generalize to novel objects and arbitrary start poses. However, NDF and its variant \cite{xue2023useek} mostly consider tabletop settings and rely on hand-crafted motion primitives.

The Deep Reinforcement Learning (DRL) methods learn to maximize a given reward for object manipulation through trail and error \cite{kalashnikov2018qt}. These methods usually learn in simulation and demonstrate sim2real generation. A recent work \cite{zhou2023hacman} utilizes actor-critic DRL for 6D object pose alignment with the given target pose through non-prehensile, i.e., pushing robot actions. Other work \cite{liu2023synergistic} integrates a vision-based RL method that performs object manipulation to eliminate the interlocked situations in constrained environment using pushing actions. In addition, the DRL are used to learn the non-prehensile robot actions for solving the tabletop rearrangement planning tasks in work \cite{yuan2018rearrangement}. Whereas in work \cite{wang2023multi}, the author purposes a multi-stage RL method for non-prehensile manipulation of objects with occluded grasping points. On the other hand, the prehensile pick-and-place actions have also been utilized within DRL for manipulation. For instance, \cite{lee2022beyond} performs the robot stacking problem of diverse shapes using the combination of DRL with a vision-based interactive policy distillation through prehensile robot actions. Others \cite{gu2016deep} propose an efficient DRL method that can handle pick-and-place actions based on off-policy training of deep Q-functions. However, there are several drawbacks to DRL-based methods. First, these approaches require a large amount of interactions with the environment to learn generalizable manipulation skills. Hence, their training process can be very time-consuming. Second, the reward design is challenging and requires significant domain knowledge \cite{teng2014self}. Lastly, these methods mostly consider tabletop environments as incorporating complex collision avoidance constraints often leads to failure or requires an even larger interaction-based dataset for learning. 

Unlike prior work, we introduce a new category, i.e., a physics-informed neural method for object manipulation. Our approach does not require expensive expert demonstrations as in imitation learning or trial and error-based learning as in DRL. Instead, our method formulates an object manipulation problem as a solution to Eikonal PDE and directly learns by solving the physics equation. We demonstrate that our data generation roughly takes half an hour. Furthermore, once our method is trained, it solves complex object manipulation tasks in cluttered environments in a fraction of a second and generalizes to novel objects without any re-training or fine-tuning. In addition, we show complex cases requiring the robots to alter grasping points to achieve the desired manipulation objectives.  

\section{Background}
This section formally presents our notations and a brief background on physics-informed NTFields and its variant P-NTFields.\vspace{-0.05in}
\subsection{Problem Defination}
Let the robot configuration space be denoted as $\mathcal{Q}$ with its obstacle and obstacle-free space indicated as $\mathcal{Q}_{obs}$ and $\mathcal{Q}_{free}$, respectively. Similarly, the environment, i.e., the robot's surrounding workspace, is denoted as $\mathcal{X}$, comprising obstacles, $\mathcal{X}_{obs} \subset \mathcal{X}$ and obstacle-free space, $\mathcal{X}_{free}= \mathcal{X}\backslash \mathcal{X}_{obs}$. The environment contains $m$ number of objects, each with a six-dimensional pose denoted as $p^i \in \mathbb{R}^6$ and $i\in [0,m]$. We consider the problem of POM, which requires a robot to manipulate the given object $i$ from its starting pose $p^i_s$ to some desired pose $p^i_g$ while avoiding collisions with the obstacles in the environment.
\subsection{Neural Time Fields}

Motion planning is a fundamental problem in robotics, where a robot must compute a collision-free trajectory between a start and goal configuration. A common approach to solving this problem is through the Eikonal equation, which describes the shortest path a wavefront travels while respecting a given speed function. The Eikonal equation inherently accounts for obstacle avoidance by assigning low speed near obstacles and high speed in free space, effectively guiding the trajectory around obstacles while ensuring efficiency in motion planning.

Traditional methods such as the Fast Marching Method (FMM) \cite{sethian1996fast, treister2016fast} numerically solve the Eikonal equation, computing the arrival time field as a wave propagates through space. While FMM provides an efficient way to compute paths, it requires space discretization, making it susceptible to the curse of dimensionality in high-dimensional planning spaces.

To address these limitations, NTFields was introduced as a PINN-based approach to solving the Eikonal equation for motion planning. Instead of relying on discrete grid-based solutions, NTFields use a neural network to implicitly represent the arrival time function $T({q}_s, {q}_g)$, which denotes the travel time for a wavefront to propagate from the start configuration ${q}_s$ to the goal configuration ${q}_g$. The method formulates motion planning as solving the Eikonal equation: \begin{equation}\label{Equation1} \frac{1}{S(q_g)} = \lVert \nabla_{q_g} T(q_s, q_g)\rVert, \end{equation} where $S(q_g)$ is a speed function that determines how the wave propagates, directly influencing the planned trajectory. By learning to approximate the solution to Eq. \ref{Equation1}, NTFields generate smooth and efficient motion plans without requiring discrete space discretization.

However, solving the Eikonal equation with neural networks is inherently challenging due to its ill-posed nature, often leading to unstable training and convergence issues. To mitigate this, P-NTFields introduced a viscosity term by incorporating a second-order Laplacian regularization $\Delta_{q_g}T(q_s, q_g)$, which stabilizes training and ensures smoother solutions. Additionally, P-NTFields incorporated a progressive learning strategy, gradually refining the learned function to improve convergence.

Both NTFields and P-NTFields use a learned speed function to shape the resulting trajectory. The ground-truth speed function is defined as: \begin{equation}\label{Equation4} S^*(q) = \frac{s_{const}}{d_{max}} \times \text{clip}(d(q, \mathcal{X}{obs}), d_{min}, d_{max}), \end{equation} where $d(q, \mathcal{X}_{obs})$ represents the minimum distance between the robot at configuration $q$ and the closest obstacle $\mathcal{X}_{obs}$. The speed function is clipped between predefined limits $d_{min}$ and $d_{max}$, then normalized and scaled by a factor $s_{const}$ to ensure numerical stability.

While P-NTFields improve training stability over NTFields, they introduce a computational overhead due to the second-order viscosity term, requiring higher-order derivatives of the neural network, which increases training time. Moreover, P-NTFields have only been applied to standard motion planning tasks, and their potential for solving POM tasks remains unexplored. In object manipulation, additional constraints arise due to object rotations, grasp constraints, and dynamic interactions with multiple objects, making the direct application of NTFields and P-NTFields to POM non-trivial.
\section{Methods}

This section presents our method for solving object manipulation tasks using PINN-based motion planning. We introduce a new regularization for solving the Eikonal equation in the object manipulation space with a modified neural architecture and an improved speed definition that considers robot space feasibility. Furthermore, we introduce the search method which views the object manipulation space travel time as a cost-to-go function, and apply search in robot configuration space to find the robot path for execution. These modifications enable kinematically feasible object manipulation in cluttered environments, scale efficiently to multi-object settings, and eliminate the need for expensive expert demonstrations.

\subsection{Dirichlet energy regularization}
While P-NTFields improve stability by introducing a viscosity term via the Laplacian operator $\Delta T$, this approach significantly increases computational overhead, as it requires computing second-order derivatives of the neural network. Our goal is to retain the regularization benefits of the viscosity-based Eikonal equation while reducing training time.

To achieve this, we propose an alternative Dirichlet Energy minimization approach. The Dirichlet energy of a function $T$ is given by:
\begin{equation}\label{Dirichlet} E_D(T) = \int_{\Omega} |\nabla T|^2 d\Omega. \end{equation} Minimizing $E_D(T)$ encourages smooth solutions while only requiring first-order derivatives, making it computationally more efficient than the viscosity-based Laplacian regularization in P-NTFields.

A key property of Dirichlet energy minimization is that, in the energy minimization framework, it is equivalent to solving the Laplace equation $ \Delta T = 0. $

Thus, by minimizing $E_D(T)$, we achieve a similar regularization effect to the viscosity term $\Delta T$, which helps stabilize training by reducing gradient fluctuations and ensuring smooth trajectory generation.

In our approach, we retain the original Eikonal equation loss but replace the viscosity term with Dirichlet energy minimization. This substitution improves training stability by reducing high-frequency artifacts in gradient updates, leading to more consistent learning dynamics. Additionally, it enables faster convergence compared to the viscosity-based formulation while maintaining smooth solutions. By avoiding the need for second-order derivatives, our approach reduces computational overhead while preserving the regularization benefits of viscosity terms.

By leveraging this Dirichlet-based formulation, our method achieves efficient and scalable object manipulation trajectory generation while ensuring robustness in complex, cluttered environments.
\subsection{Time Field Generator}
Unlike motion planning in robot configuration space, our approach focuses on object manipulation in pose space, where each object's position and orientation are explicitly considered. Instead of planning in terms of robot joint space, we directly formulate the problem in object pose space, enabling kinematically feasible object movements. We use $p$ to represent an object’s pose and $o$ to denote the object index in multi-object scenarios.

The time field generator, denoted as $g()$, predicts the time field $T(p^o_s,p^o_g)$. The input to this model is the start and goal pose of a given object and the output is time $T$, i.e., 
\begin{equation}\label{Equation10}
    T(p^o_s, p^o_g) = g\big([(f(p^o_s) \bigotimes f(p^o_g)) , k(o)]\big)
\end{equation}
where $f$, $\bigotimes$, and $k$ denote the object pose encoder, symmetric operator, and object shape encoder, respectively. The pose encoder, $f$, passes the given pose through a random Fourier feature \cite{rahimi2007random, tancik2020fourier} to obtain high-frequency object pose features. These features are then further processed by a Multi-Layer Perceptron (MLP) to obtain the latent embeddings of the object's given poses. The $\bigotimes$ is a symmetric operator that concatenates $\max$ and $\min$ of the latent embeddings of start and goal poses. This operator is inspired by NTField and P-NTFields, which enforces the symmetry property of the Eikonal Equation. That is, the arrival time from start to goal and from goal to start should be the same. Finally, the symmetrized latent embedding of object poses is concatenated with the object shape embeddings. These are then passed through function $g$ to predict $T$.

The shape encoder $k$ enables our time field generator to generalize to different objects and generate their manipulation trajectories. Recall in Eq. \ref{Equation4}; the ground truth speed is calculated based on the minimal distance $d$ to the obstacles of the given configuration. Therefore, aside from the pose, the shape and scale of the objects embedded in their point clouds are also the key factors in deciding $d$. Thus, we utilize the PointNet++ \cite{qi2017pointnet++} structure denoted as $k()$ to encode object $o$ and generate the latent embedding of various objects from their point clouds. The Single-Scale Grouping (SSG) is used internally in PointNet++ to capture the features of the objects. This encoder also allows our method to generalize to never-seen-before objects as it captures the relationship between the minimal distance toward the obstacles and the object's shape and scale. The function $g$ is a ResNET-style MLP with skip connections. 

Given the predicted time field $T(p^o_s, p^o_g)$, the speed field is estimated using Equation \ref{Equation1}.

\subsection{Improved Speed Model}
The P-NTFields and NTFields were designed for robot motion planning. Their ground truth speed models are inapplicable to solving POM tasks as they do not take into account objects, their poses, and the robot's reachability. Therefore, we introduce the following new formulation for the ground truth speed function $S^*(p^o)$, which leads to kinematically feasible object manipulation.
\begin{equation}\label{Equation11}
    \begin{aligned}
         S^*(p^o) = \frac{s_{const}}{d_{max}} \times \text{clip}(&d(X_{p^o}, \mathcal{X}_{obs})\times \text{IK}_r(p^o), \\&~d_{min}, d_{max}),
    \end{aligned}
\end{equation}
where $X_{p^o}$ denotes the point cloud of object $o$ when its pose is $p^o$. In addition, we also check if the center location of the object at each sampled pose is kinematically reachable by the robot. It is achieved by the indicator function $\text{IK}_r(p^o)$ in the above equation. If the object pose is not reachable, the ground truth speed is set to be minimal, indicating an invalid region for object manipulation.

\subsection{Training pipeline}

Our neural network, defined in Eq. \ref{Equation10}, predicts the arrival time field $T$ and its gradients with respect to the start and goal poses. These gradients are then used to estimate the speed field $S$. To ensure the predicted speed matches the ground truth speed (Eq. \ref{Equation11}), we employ an isotropic loss function:

\begin{equation}\label{Equation12}
    \begin{aligned}
        L(S^*(p^o), S(p^o)) = & S^*(p^o_s)/S(p^o_s) +  S(p^o_s)/S^*(p^o_s) + \\
        & S^*(p^o_g)/S(p^o_g) + S(p^o_g)/S^*(p^o_g) - 4
    \end{aligned}
\end{equation}

This loss function, originally introduced in NTFields, is smooth and lower-bounded by zero, providing an optimization objective. We optimize this loss with Dirichlet energy (Eq. \ref{Dirichlet}) and adopt the progressive speed scheduling from P-NTFields, allowing the PINN model to effectively learn collision-free, kinematically feasible object manipulation trajectories.

The final training objective is formulated as:

\begin{equation}\label{loss} \begin{aligned} \min_{\theta} \int_{\Omega} L(S^*(p^o), S(p^o)) + \epsilon |\nabla_{p_g} T(p_s,p_g)|^2. \end{aligned} \end{equation}

This formulation ensures that the network learns an accurate and stable time field by minimizing the discrepancy between the predicted and ground truth speed while incorporating Dirichlet energy minimization for gradient regularization. By leveraging these techniques, our PINN-based approach efficiently learns object manipulation trajectories, ensuring fast convergence and robust generalization to unseen environments and objects.
\subsection{Multimodal Manipulation Planning with PINN}
During the testing time, the neural network predicts $T$  (Equation \ref{Equation10}) for the given object and its start and goal poses, $(p^o_s,p^o_g)$. Then, the arrival time $T(p^o_s, p^o_g)$ and the speed fields at start and goal poses $S(p^o_s), S(p^o_g)$ are computed using Equation \ref{Equation1}, respectively. The object motion trajectory, denoted as $\sigma_o$, is generated in a bidirectional manner by growing the start pose and the goal pose towards each other as follows:\\
\begin{equation}\label{Equation13}
    \begin{aligned}
        &p^o_s \prime \leftarrow p^o_s - \eta S^2(p^o_s) \nabla_{p^o_s}T(p^o_s, p^o_g)\\
        &p^o_g \prime \leftarrow p^o_g - \eta S^2(p^o_g) \nabla_{p^o_g}T(p^o_s, p^o_g)
    \end{aligned}
\end{equation}
The $\eta$ controls the step size about how fast they move to each other. The above-mentioned step is performed iteratively until the distance between the two poses $\lVert p^o_s \prime-p^o_g \prime \rVert_2$ is smaller than a tiny threshold $d_s$. The trajectory $\sigma_o$ stores all the intermediate poses between the given start and goal.
\begin{algorithm}[h!]
\caption{Object Manipulation (OManip)}\label{alg:robot_traj}
\textbf{Inputs:} POM-NeTF$(\cdot)$, $o, p^o_s, p^o_g, \text{validgrasps}$\;
$\sigma_o =$ POM-NeTF$(o, p^o_s, p^o_g)$ \tcp*[l]{object motion}
\For{G $\in$ \text{validgrasps}}{
    $\sigma_r = IK(\sigma_o, G)$ \; \tcp*[l]{robot configuration trajectory}
    \If {not $\sigma_r$}{
        $p^o_c = \text{decouple}(p^o_s, p^o_g)$ \tcp*[l]{intermediate pose}
        $\sigma_r^1 = \text{OManip}(o, p^o_s, p^o_c)$\tcp*[l]{recursion}
        $\sigma_r^2 = \text{OManip}(o, p^o_c, p^o_g)$ \tcp*[l]{recursion}
        \If {$\sigma_r^1$ and $\sigma_r^2$}{
            return $\text{RMP}(\sigma_r^1 + \sigma_r^2)$
        }
    }
    \Else{
        return $\text{RMP}(\sigma_r)$
    }
} 
\end{algorithm}
Next, we describe our multi-step object manipulation approach that leverages the POM-NeTF within the loop to generate kinematically feasible robot motion sequences for execution. Algorithm~\ref{alg:robot_traj} summarizes our strategy named OManip. The inputs to the algorithm are the trained POM-NeTF, robot valid grasps, the object $o$, and its start/goal poses $p^o_s, p^o_g$ (Line 1). In our implementation, we use Contact-GraspNet \cite{sundermeyer2021contact} as our grasp generation function. This function provides valid robot grasps for the given object sorted by their stability scores. The algorithm begins by generating the object trajectory $\sigma_o$ using POM-NeTF (Line 2).  Next, it searches through valid grasps and uses the inverse kinematics function to determine if all object poses $(\sigma_o)$ can be mapped into valid robot configurations $\sigma_r$ (Lines 3-4). If the mapping is successful, the resulting robot configuration sequences are further smoothed out by the Riemannian Motion Policy (RMP) \cite{ratliff2018riemannian} for execution (Line 12). However, if the mapping fails, a new intermediate pose $p^o_c$ between the start and goal is proposed. This intermediate pose is selected to perform in-place manipulation along one dimension of rotation. This pose is generated by our decouple function (Line 6). The dimension of rotation is selected to be the one farther away from its goal. In practice, we find out that creating intermediate pose $p^o_c$ by separating in-place manipulation and remaining manipulation leads to better search efficiency. Once we get the intermediate pose $p_o^c$ that divides the original task into two sub-tasks, we recursively execute the Omanip function on both of them (Lines 7-8). If the solutions for both sub-tasks are found, they are combined as the final robot configuration trajectory involves changing the grasping position (Lines 9-10). Note that as we call OManip recursively, it searches over the given grasp again, this allows for our method to change grasps during execution, leading to robust, multimodal POM. Furthermore, since we trained our POM-NeTF to respect the robot's kinematic reachability (Equation 11), the OManip mostly converges very quickly, as evident from our results. 

\vspace{-0.2cm}
\begin{figure}[htp]
\centering
\includegraphics[width=.15\textwidth]{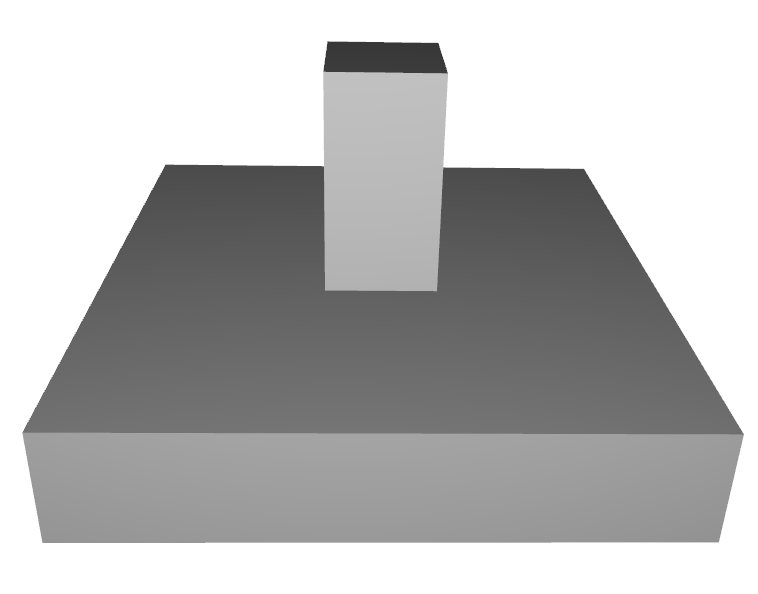}\hfill
\includegraphics[width=.13\textwidth]{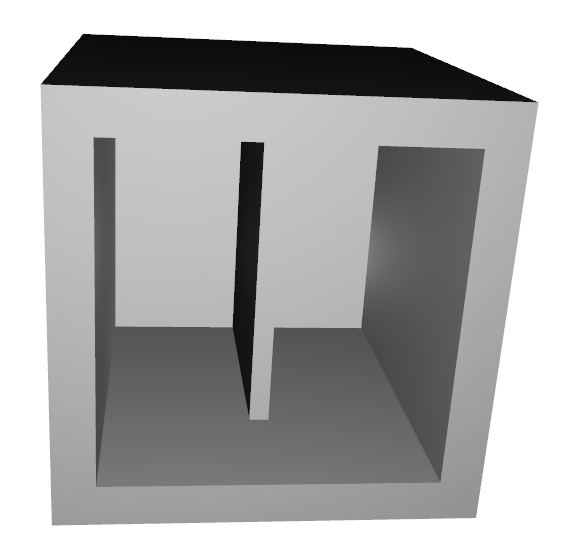}\hfill
\includegraphics[width=.18\textwidth]{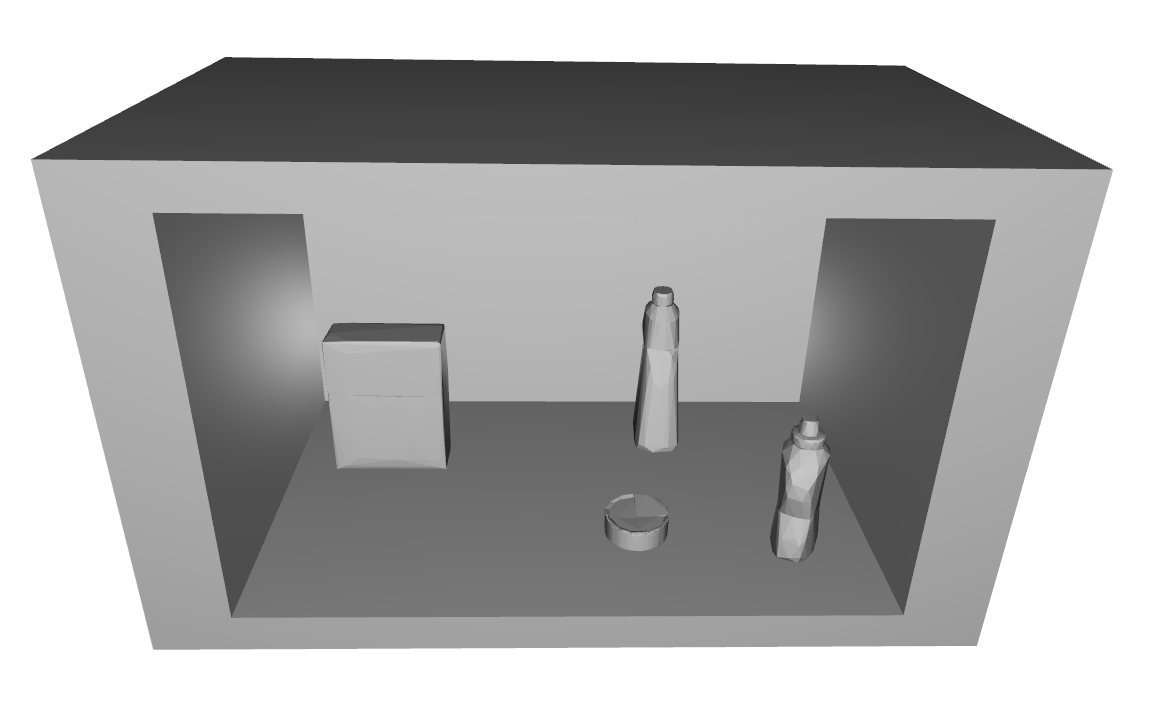}
\caption{Depiction of our three environment setups: (Left) the tabletop with a center obstacle, (Middle) the U-shaped tunnel, and (Left) the cabinet.}
\label{fig:environments}
\vspace{-0.6cm}
\end{figure}
\subsection{Implementation Details}
We create three constrained environments shown in Fig. \ref{fig:environments} to train and test our method, i.e., a tabletop environment of size 0.5 m$^3$ with a huge obstacle in the center, a U-shaped confined tunnel of size 1 m$^3$, and a real-world cabinet-like environment cluttered with household objects. The cabinet is a confined environment with the following $xyz$ dimensions $(0.55~\text{m}, 1~\text{m}, 0.6~\text{m})$. For the objects, we include 50 objects from the YCB dataset \cite{calli2015benchmarking}. 45 of them are chosen to be involved in the training, while the remaining 5 are preserved as never-seen-before objects that only appear during the testing time to verify the network's generalization capability. For each object in a certain environment, 50k valid tuples are collected with each of them composed of the object identifier $o$, the start/goal poses $p^o_s, p^o_g \in \mathbb{R}^6$, and the corresponding ground truth speed $S^*(p^o_s), S^*(p^o_g)$ that is determined by Eq. \ref{Equation11}.
The point clouds of all objects are converted to the same size of 64 points using the farthest point down-sampling method \cite{moenning2003fast}. The object identifier $o$ is stored as an integer that maps to its name in a pre-configured file, which further links it to the corresponding point cloud. The data generation for the three environments takes 1893.156 s, 2047.588 s, and 3249.305 s, respectively. We implement our method using Pytorch. The code repository will be open-sourced with the final manuscript.

\begin{table*}[!ht]
\fontsize{7}{5}\selectfont
  \begin{center}
    \begin{tabular}{c c c c c c c c c c}
      \hline
      \multirow{2}{*}{\textbf{Method}} & \multicolumn{3}{c}{\textbf{Table w/ obs}} & \multicolumn{3}{c}{\textbf{U-shaped Tunnel}} & \multicolumn{3}{c}{\textbf{ Cabinet}} \Tstrut\Bstrut \\
       & Time (s) $\downarrow$ \Tstrut\Bstrut & Length (m) $\downarrow$ & SR ($\%$) $\uparrow$ & Time (s) $\downarrow$ \Tstrut\Bstrut & Length (m)$\downarrow$ & SR ($\%$) $\uparrow$ & Time (s) $\downarrow$ \Tstrut\Bstrut & Length (m) $\downarrow$ & SR ($\%$) $\uparrow$\\
       \hline
       Ours \Tstrut\Bstrut & \textbf{0.065 $\pm$ 0.009} & 1.153 $\pm$ 0.121 & 98.8 & \textbf{0.065 $\pm$ 0.053} & \textbf{1.704 $\pm$ 1.683} & 94.4 & \textbf{0.048 $\pm$ 0.014} & \textbf{0.819 $\pm$ 0.227} & 98.8\\
       P-NTFields \Tstrut\Bstrut & 0.066 $\pm$ 0.008 & \textbf{1.144 $\pm$ 0.115} & 98.4 & 0.067 $\pm$ 0.024 & 1.789 $\pm$ 1.712 & 94.2 & 0.052 $\pm$ 0.015 & 0.831 $\pm$ 0.227 & 99.2\\
       NTFields \Tstrut\Bstrut & 0.074 $\pm$ 0.012 & 1.196 $\pm$ 0.124 & 96.8 & 0.074 $\pm$ 0.095 & 1.811 $\pm$ 2.046 & 87.6 & 0.062 $\pm$ 0.018 & 0.913 $\pm$ 0.308 & 99.2\\
       RRT-Connect \Tstrut\Bstrut & 0.340 $\pm$ 0.720 & 2.961 $\pm$ 1.813 & 98.4 & 0.821 $\pm$ 3.121 & 3.537 $\pm$ 3.213 & 92 & 0.153 $\pm$ 0.170 & 1.420 $\pm$ 0.408 & 97.2\\ 
       JIST-Based \Tstrut\Bstrut & 0.530 $\pm$ 0.944 & 3.360 $\pm$ 1.945 & \textbf{100} & 1.206 $\pm$ 4.186 & 3.819 $\pm$ 3.452 & \textbf{97.2} & 0.235 $\pm$ 2.947 & 1.503 $\pm$ 0.546 & \textbf{100}\\ 
       \hline
       \end{tabular}
       \caption{Across all testing environments, our POM-NeTF achieve the best overall performance in the planning time and the trajectory length while maintaining high success rates.}
    \label{tab:table_comp}
  \end{center}
    \vspace{-0.1in}
    \end{table*}

\begin{figure*}[ht]
\vspace{-0.04in}
    \centering
    \includegraphics[trim = {0cm 0cm 0cm 0cm}, clip, width=1\textwidth]{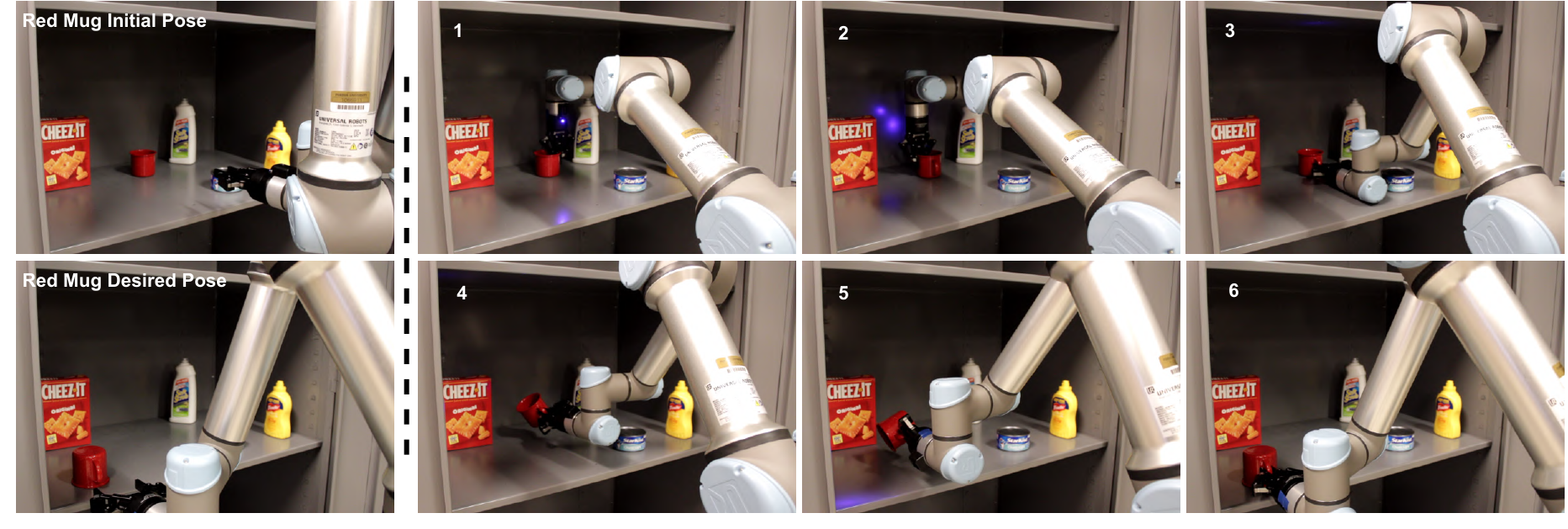}
    \caption{Demonstration of POM-NeTF enabling the robot to manipulate the red mug to the desired pose. The figures on the left depict start and goal poses. It can be seen that our method effectively plans object motion trajectory and autonomously determines the intermediate placement pose to allow the robot to reposition its grasp to achieve the remainder of the manipulation objective. In this scenario, the intermediate pose requires the robot to grasp from the rim (Frame 1), perform in-place manipulation to rotate the mug (Frame 2), and then regrasp from its handle to move to the target pose.}
    \label{fig:exp1-figure}
    \vspace{-0.3cm}
\end{figure*} 

\section{Results}
In this section, we first present the results and analysis of our method against the state-of-the-art baselines in the simulation and real-world environment. These experiments were performed on a system equipped with a 3.60GHz $\times$ 16 Intel Core i7 processor with 32GB RAM and a Nvidia GeForce RTX 3090 GPU. Furthermore, we consider the following baselines and metrics for comparison.
\begin{itemize}[leftmargin=*]
    \item \textbf{RRT-Connect \cite{kuffner2000rrt}}: This is a sampling-based method to connect the start and end poses in a bidirectional manner. This results in a trajectory that is followed by also sampling the suitable robot grasps required for POM. In order to fairly compare it with our method, the objective of the baseline is set to find the minimum distance while maximizing the minimal distance of the object from the obstacles. 
    \item \textbf{JIST-based \cite{kimmel2018fast}}: This method is based on the Jacobian Informed Search Tree (JIST). It samples gripper placement and also uses a search strategy similar to RRT-Connect to find trajectories that minimize movement distance and maximize safety margins from obstacles.
    \item \textbf{NTFields}: This baseline is inspired by the original NTFields framework \cite{ni2022ntfields}. It uses the Eikonal equation \ref{Equation1}. 
    \item \textbf{P-NTFields}: This baseline is inspired by the original P-NTFields that adds the viscosity term in the Eikonal equation \ref{Equation1} and uses the progressive learning method to solve the inaccurate time field prediction problem around the obstacles. 
\end{itemize}
Please note that the original NTFields and P-NTFields methods cannot solve POM tasks. We augment these baselines with our training pipelines, loss function, neural architecture, and planning procedures to enable them to solve POM tasks. The P-NTFields use computationally expansive viscosity terms based on Laplacian and can be viewed as an expert since our method attempts to approximate viscosity via computationally cheaper Dirichlet energy minimization.  
In the case of sampling-based baselines, to enforce fair comparison with our methods, they also construct object trajectories in the workspace and then use our OManip for robot motion generation. We use the Open Motion Planning Library (OMPL) \cite{sucan2012open} and Flexible Collision Library (FCL) \cite{pan2012fcl} for the classical baseline implementation. \par
To compare the performance of the above-mentioned methods, we use the following evaluation metrics:
\begin{itemize}[leftmargin=*]
    \item \textbf{Planning time (Time)}: It measures the total time used to generate a feasible object manipulation plan that transfers the object from the start pose to the goal pose. All methods construct the object trajectories first and then leverage the proposed OManip framework to generate the corresponding robot trajectories.
    \item \textbf{Length}: It records the object's total moving Euclidean distance by following a  given trajectory.
    \item \textbf{Success rate (SR)}: It reflects the percentage of feasible collision-free object manipulation trajectory.
    \item \textbf{Training time}: For all PINN-based methods, we also report their training times in hours to highlight the significance of Dirichlet energy minimization over the viscosity term. 
\end{itemize}
\subsection{Simulation Results}
The simulation experiments are conducted in the three environments described in Section IV-F. In each environment, we create 500 test cases with YCB objects, making a total of 1500 test cases. 

Table \ref{tab:table_comp} presents the results for three simulation environments. It can be seen that overall, our method achieves the best performance in terms of the planning times and path lengths. In the tabletop environment, P-NTFields has slightly better path lengths than ours. Additionally, in terms of success rate in the cabinet environment, P-NTFields also outperformed our method by a small margin of 0.4\%. This suggests that Dirichlet energy minimization successfully approximated the viscosity. Although the viscosity can still perform slightly better, this comes at a significant computational cost, as highlighted in Table \ref{tab:time}. The training times of P-NTFields are almost double, as it requires the calculation of a double derivative viscosity term. In contrast, our approach is more training efficient as it approximates viscosity with Dirichlet energy minimization and requires only a single derivative. When comparing our method against NTFields, it can be seen that our approach takes similar training times but always finds shorter trajectories, exhibits higher success rates, and lower planning times, which further validates the effectiveness of Dirichlet energy minimization.  In the case of the sampling-based methods, their planning times are high, and trajectory length is still at least 50 $\%$ worse than our method, even with the path smoothing. Although there exist optimal sampling-based planners such as RRT* \cite{karaman2011sampling}, we exclude them from comparison due to their very high computational times.
\begin{table}[h!]
\begin{center}
\resizebox{0.48\textwidth}{!}{%
\begin{tabular}{ c  c c c  } 
\hline
\textbf{Method} & Table w/ obs & U-shaped tunnel & Cabinet\\
\hline
Ours & 1.29  & 1.33  & 1.38  \\
P-NTFields & 2.73  & 2.59  & 2.96 \\
NTFields & 1.30  & 1.31  & 1.34 \\
\hline
\end{tabular}}
\end{center}
\caption{\label{tab:time}Training time (hours) for POM-NeTF, P-NTFields, and NTFields}
\vspace{-0.3cm}
\end{table}
\begin{figure}[ht]
\vspace{-0.04in}
    \centering
    \includegraphics[trim = {0cm 0cm 0cm 0cm}, clip, width=0.46\textwidth]{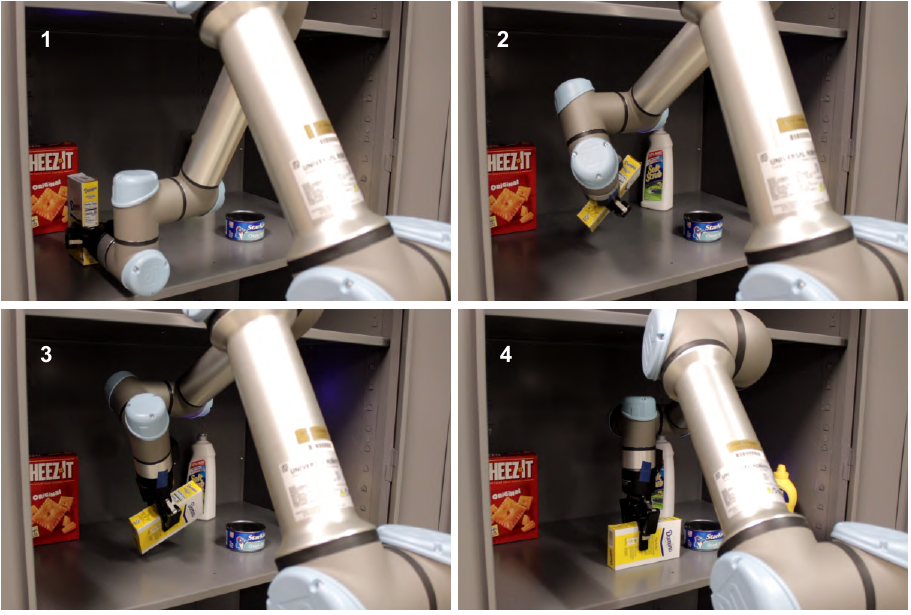}
    \caption{POM-NeTF manipulates an unseen sugar box. Starting and goal positions in Frames 1 and 4 require major rotation adjustments. The robot arm smoothly executes the trajectory from POM-NeTF in one pass.}
    \label{fig:exp2-figure}
    \vspace{-0.3cm}
\end{figure}

\subsection{Real Robot Experiments}
We conduct real robot experiments to examine the sim2real capability of our method and compare it against the P-NTFields and the RRT-Connect. We exclude NTFields and JIST planners due to their inferior performance in the simulated environments. Our testing environment is a confined cabinet of size $($0.55 m, 1 m, 0.6 m$)$. In this cabinet, we place three seen and two unseen objects. For each object, two start/goal pairs that involve major changes in translation and orientation are created. For these pairs, the above-mentioned planners are executed to find the POM trajectories. Our results are summarized in Table \ref{table:real-world comp}.  It can be seen that our method performs similarly to P-NTFields. These physics-informed methods outperform RRT-connect in all metrics by a significant margin.

Fig. \ref{fig:exp1-figure} and \ref{fig:exp2-figure} demonstrate our method's execution instances in the real world for manipulating a mug and a sugar box, respectively. In the former case, the robot grasped the mug to reorient it and then re-grasped it from the handle to reach the target configuration. In the latter case, our method reached the target sugar box pose without re-grasping. These cases highlight that our method can perform almost real-time multimodal POM, where the robot can re-grasp objects whenever needed and generate their manipulation trajectories for reaching the complex target poses. We also provide additional videos of our real-world experiments in the supplementary material.

\begin{table}[!ht]
  \begin{center}
      \begin{tabular}{c c c c}
      \hline
           Method \Tstrut\Bstrut & Time (s) $\downarrow$ & Length (m) $\downarrow$& SR $(\%)\downarrow$\\
           \hline
          Ours \Tstrut\Bstrut & $0.062 \pm 0.017$ & $1.021 \pm 0.191$ & $100$\\
          P-NTFields \Tstrut\Bstrut & $0.062 \pm 0.016$ & $1.083 \pm 0.213$ & $100$\\
          RRT-Connect \Tstrut\Bstrut & $0.735\pm0.484$ & $3.689\pm1.871$ & $100$\\
          \hline
      \end{tabular}
      \caption{Results of various methods in solving multimodal POM in real-world confined, cabinet environment.}
      \label{table:real-world comp}
  \end{center}
      \vspace{-0.3in}
\end{table}

\section{Conclusion}
In this paper, we present a new approach to solving object manipulation problems using physics-informed neural networks (PINNs) called POM-NeTF. Our method can find object manipulation trajectories without the need for expert demonstration by directly solving the Eikonal equation. We introduce a computationally efficient strategy based on Dirichlet energy minimization, reducing training time while maintaining high performance. To handle object manipulation trajectories that are impossible to achieve with one robot grasp, our method performs multimodal contact and manipulation planning, leading to the successful execution of complex object motion trajectories. Our future work includes adding kinematic constraints to our object manipulation framework to solve more articulated tasks such as tool use in daily life environments.

\bibliographystyle{unsrt}
\bibliography{root}
\end{document}